\documentclass[letterpaper, 10 pt, journal, twoside]{IEEEtran}  

\usepackage{amssymb,amsmath,amsfonts,mathrsfs}
\usepackage{graphicx}
\usepackage[table,xcdraw,dvipsnames]{xcolor}
\usepackage{tikz}
\usepackage{url}

\usepackage[labelformat=simple]{subcaption}

\usepackage{colortbl}
\usepackage{booktabs}
\usepackage{array}
\usepackage{tabularx}

\usepackage[normalem]{ulem}

\usepackage[flushleft]{threeparttable}
\usepackage{multirow}

\let\labelindent\relax
\usepackage{enumitem}

\newcolumntype{L}[1]{>{\raggedright\let\newline\\\arraybackslash\hspace{0pt}}m{#1}}

\makeatletter
\DeclareRobustCommand\onedot{\futurelet\@let@token\@onedot}
\def\@onedot{\ifx\@let@token.\else.\null\fi\xspace}

\usepackage[T1]{fontenc}  
\usepackage[b]{esvect}    

\makeatletter
\newlength\xvec@height%
\newlength\xvec@depth%
\newlength\xvec@width%
\newcommand{\xvec}[2][]{%
  \ifmmode%
    \settoheight{\xvec@height}{$#2$}%
    \settodepth{\xvec@depth}{$#2$}%
    \settowidth{\xvec@width}{$#2$}%
  \else%
    \settoheight{\xvec@height}{#2}%
    \settodepth{\xvec@depth}{#2}%
    \settowidth{\xvec@width}{#2}%
  \fi%
  \def\xvec@arg{#1}%
  \def\xvec@dd{:}%
  \def\xvec@d{.}%
  \raisebox{.2ex}{\raisebox{\xvec@height}{\rlap{%
    \kern.05em
    \begin{tikzpicture}[scale=1]
    \pgfsetroundcap
    \draw (.05em,0)--(\xvec@width-.05em,0);
    \draw (\xvec@width-.05em,0)--(\xvec@width-.15em, .075em);
    \draw (\xvec@width-.05em,0)--(\xvec@width-.15em,-.075em);
    \ifx\xvec@arg\xvec@d%
      \fill(\xvec@width*.45,.5ex) circle (.5pt);%
    \else\ifx\xvec@arg\xvec@dd%
      \fill(\xvec@width*.30,.5ex) circle (.5pt);%
      \fill(\xvec@width*.65,.5ex) circle (.5pt);%
    \fi\fi%
    \end{tikzpicture}%
  }}}%
  #2%
}
\makeatother

\makeatletter
\renewcommand*\env@matrix[1][\arraystretch]{%
  \edef\arraystretch{#1}%
  \hskip -\arraycolsep
  \let\@ifnextchar\new@ifnextchar
  \array{*\c@MaxMatrixCols c}}
\makeatother

\definecolor{commentcolor}{gray}{0.5}
\usepackage{algorithm}
\usepackage{algpseudocode}
\algrenewcommand\algorithmicindent{1.0em}%

\algnewcommand{\LineComment}[1]{\State \textcolor{commentcolor}{\(\triangleright\) #1}}
\algnewcommand{\NewComment}[1]{\textcolor{commentcolor}{\(\triangleright\) #1}}
\algnewcommand{\To}{\textbf{to}}
\algnewcommand{\Break}{\textbf{break}}
\algnewcommand{\Continue}{\textbf{continue}}
\algnewcommand{\IIf}[1]{\State\algorithmicif\ #1\ \algorithmicthen}
\algnewcommand{\EndIIf}{\unskip}
\algnewcommand{\var}[1]{\textit{#1}}
\algnewcommand{\func}[1]{\textsc{#1}}

\begin{document}

\title{Deformable One-Dimensional Object Detection\\for Routing and Manipulation}

\author{Azarakhsh Keipour$^{1}$, 
Maryam Bandari$^{2}$ 
and Stefan Schaal$^{3}$
\thanks{Manuscript received: September 9, 2021; Revised December 9, 2021; Accepted January 10, 2021.}%
\thanks{This paper was recommended for publication by Editor Cesar Cadena Lerma upon evaluation of the Associate Editor and Reviewers’ comments.}%
\thanks{This work was supported by X, The Moonshot Factory residency program.}
\thanks{$^{1}$ Azarakhsh Keipour is with The Robotics Institute, Carnegie Mellon University, Pittsburgh, PA
        {\tt\small keipour@cmu.edu}}%
\thanks{$^{2,3}$ Maryam Bandari and Stefan Schaal are with Google, Mountain View, CA {\tt\small [maryamb, sschaal] @google.com}}%
\thanks{Digital Object Identifier (DOI): see top of this page.}%
}%

\markboth{IEEE Robotics and Automation Letters. Preprint Version. Accepted, January 2022}{Keipour \MakeLowercase{\textit{et al.}}: Deformable One-Dimensional Object Detection}

\maketitle

\begin{abstract}

Many methods exist to model and track deformable one-dimensional objects (e.g., cables, ropes, and threads) across a stream of video frames. However, these methods depend on the existence of some initial conditions. To the best of our knowledge, the topic of detection methods that can extract those initial conditions in non-trivial situations has hardly been addressed. The lack of detection methods limits the use of the tracking methods in real-world applications and is a bottleneck for fully autonomous applications that work with these objects.

This paper proposes an approach for detecting deformable one-dimensional objects which can handle crossings and occlusions. It can be used for tasks such as routing and manipulation and automatically provides the initialization required by the tracking methods. Our algorithm takes an image containing a deformable object and outputs a chain of fixed-length cylindrical segments connected with passive spherical joints. The chain follows the natural behavior of the deformable object and fills the gaps and occlusions in the original image. Our tests and experiments have shown that the method can correctly detect deformable one-dimensional objects in various complex conditions.

\end{abstract}

\begin{IEEEkeywords}
Object Detection, Segmentation and Categorization; Perception for Grasping and Manipulation; Computer Vision for Medical Robotics; Computer Vision for Automation; Deformable Object Detection.
\end{IEEEkeywords}

\IEEEpeerreviewmaketitle

\section{Introduction} \label{sec:intro}

\IEEEPARstart{M}{anipulating} non-rigid objects using robots has long been the subject of research in various contexts~\cite{frobt.2020.00082}. A specific class of non-rigid objects is called Deformable One-dimensional Objects (DOOs) or Deformable Linear Objects (DLOs) and includes objects such as ropes, cables, threads, sutures, and wires. 

In order to achieve full autonomy in applications working with DOOs, one of the essential and challenging parts is perception. Many applications require complete knowledge of the object's initial conditions, even in the occluded parts. Moreover, routing and manipulation applications also require a representation model of the DOO that allows simulation and the computation of its dynamics.

There are many methods proposed in the medical imaging community that can find the DOOs (known as tubular structures in that context) in the image frame~\cite{KRISSIAN2000130, 28493, Wang2021, Noble2011, 32226}. While these methods are perfected for low signal-to-noise images, and some can even work with self-crossings, they mainly only provide the region in the image containing the DOO and are considered segmentation methods.

Various algorithms have been proposed to track a DOO across the video frames, even in the presence of occlusions and self-crossings. These methods may devise different tools such as registration methods (e.g., Coherent Point Drift~\cite{5432191}), learning, dynamics models, and simulation to predict and correct the prediction across the consecutive frames~\cite{5980431, 15711, wang2020tracking, rastegarpanah, 6630714, 8560497, af66d77e53304e6bbb64049bb46193cb, 8206058}. While some of these methods can initialize the DOO in the first frame using trivial conditions (e.g., a straight rope in camera view), the other methods require even a simple DOO configuration to be provided to them a priori. Most tracking methods will fail to initialize in even slightly complex initial conditions. 

The initial conditions are not generally provided in real-world applications, and pure segmentation is insufficient. For example, an aerial manipulator working with electrical wires at the top of a utility pole needs to detect the desired wire, including its occluded parts and crossings, and requires a model of the detected wire to perform any task on it.

\begin{figure}[!t]
\centering
    \begin{subfigure}[b]{0.235\textwidth}
        \includegraphics[width=\textwidth, height=2.3cm]{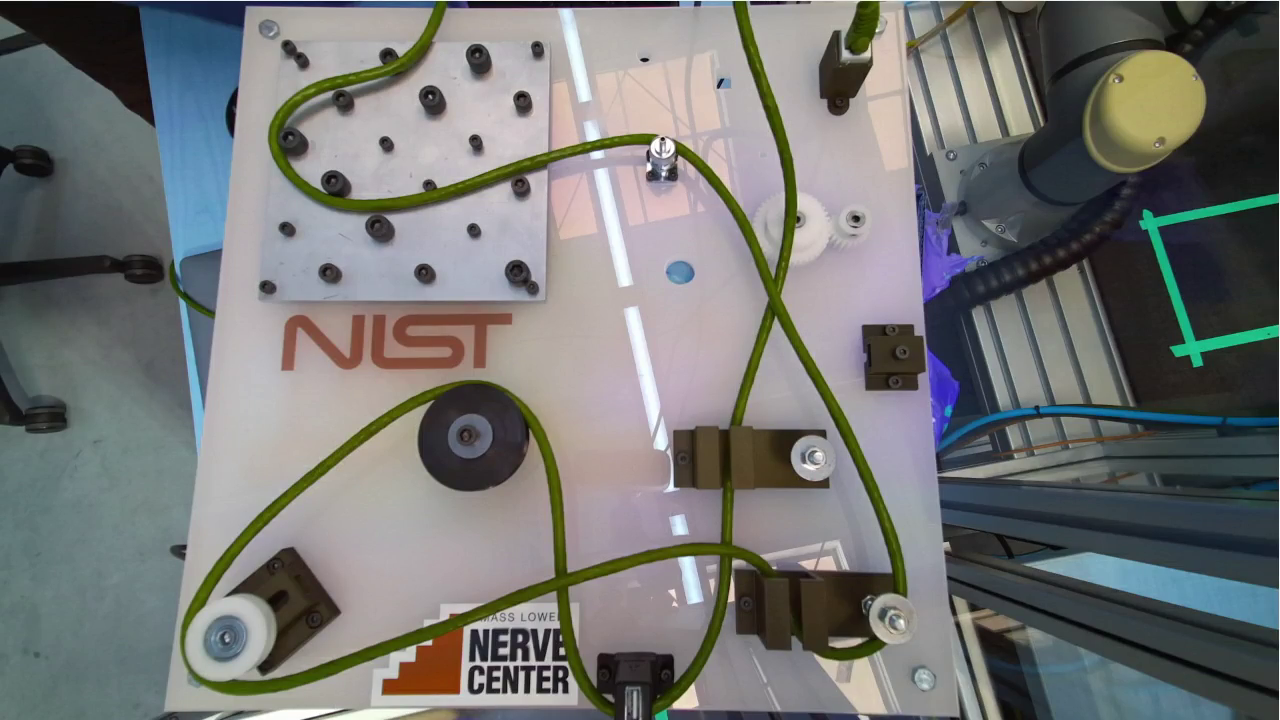}
    \end{subfigure}
    \hfill
    \begin{subfigure}[b]{0.235\textwidth}
        \includegraphics[width=\textwidth, height=2.3cm]{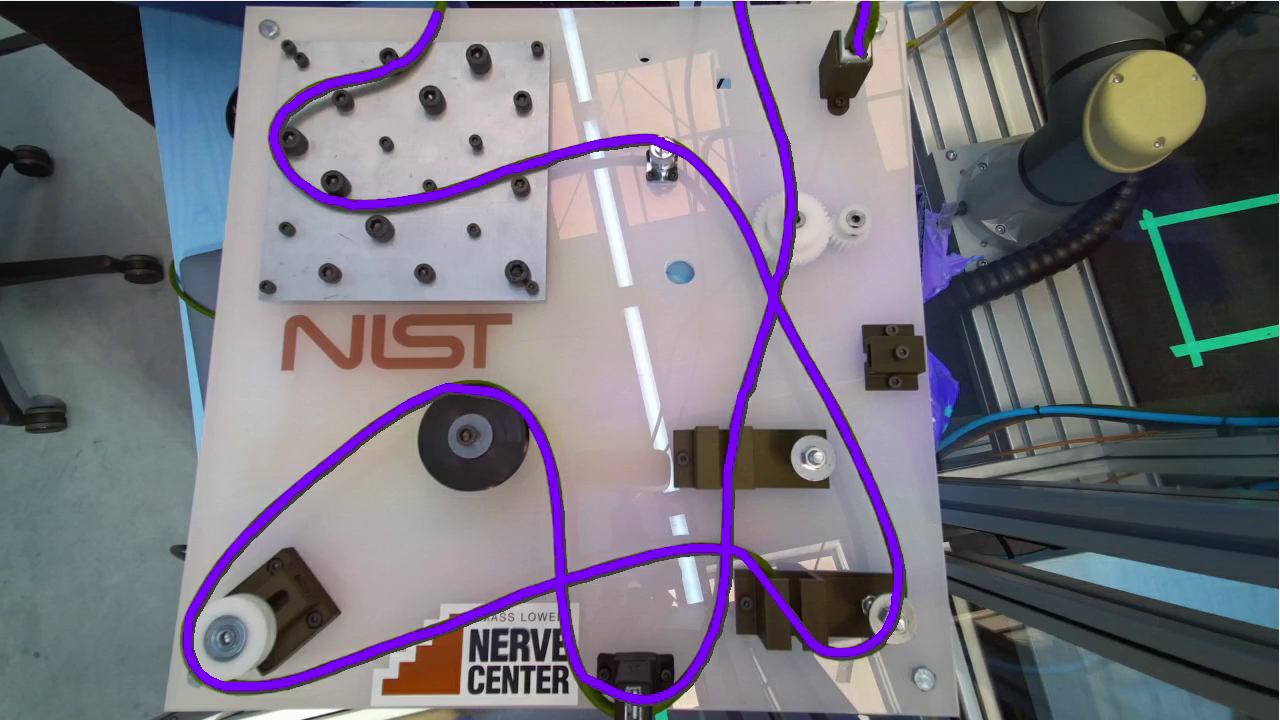}
    \end{subfigure}
    
    \medskip
    \begin{subfigure}[b]{0.235\textwidth}
        \includegraphics[width=\textwidth, height=2.5cm]{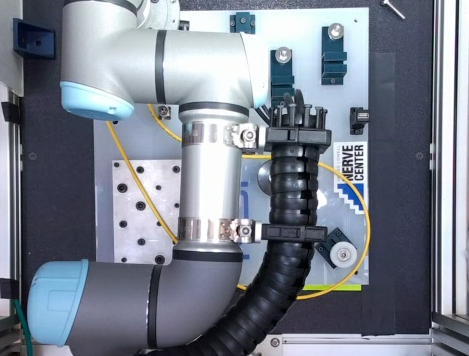}
    \end{subfigure}
    \hfill
    \begin{subfigure}[b]{0.235\textwidth}
        \includegraphics[width=\textwidth, height=2.5cm]{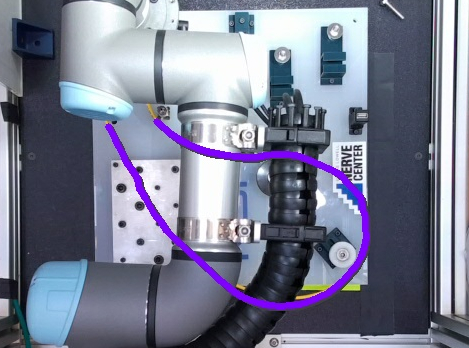}
    \end{subfigure}    
    \caption{The result of the proposed detection method on example inputs with occlusions and crossings. The detected cable (purple) overlaid on the frames on the right.}
\label{fig:results-1}
\end{figure}

The authors of \cite{8972568} proposed a method that trains on rendered images and fine-tunes on real images to detect and track the state of a rope. The method requires tens of thousands of rendered and real images of the same rope for training and needs re-training and tuning for a new DOO. This approach may be acceptable for an industrial application focusing on DOOs of the exact same characteristics, but it may not be practical for many other applications.

This paper proposes a method to detect the initial conditions of a deformable one-dimensional object. The method fills the occluded parts and works with self-crossings. The output is a single object represented as a discretized model useful for routing, manipulation, and simulation. The detected object can be used to initialize other existing tracking methods to be used in applications such as manipulating DOOs into desired shapes and knots~\cite{9044123, 8403315, 0278364906064819, 4359263}.

Sections~\ref{sec:problem} and~\ref{sec:method} define the problem and present the proposed method for detection of DOOs; Section~\ref{sec:tests} describes our implementation and shows the experiments and results. Finally, Section~\ref{sec:conclusion} discusses how the proposed method can be further improved in the future.

\section{Problem Definition} \label{sec:problem}

This work addresses the detection of cable-like deformable shapes. Unlike rigid objects, the shape of deformable objects can change, and some form of a flexible model is required to represent the current state of their shape. On the other hand, to predict the reaction of the deformable objects to the applied forces and moments, the representation model should facilitate the integration of a dynamics model. 

In theory, a DOO represented by its pixels (or voxels) in the camera frame can be integrated with a dynamics model. However, the model would typically require finite element analysis and is computationally expensive, making it impractical for robotics applications. Simpler representations are commonly used in tasks such as manipulation, routing, and planning. Arriola-Rios et al.~\cite{frobt.2020.00082} provide an overview of the common representations for deformable objects.

\begin{figure}[t]
    \centering
    \includegraphics[width=0.6\linewidth, height=1cm]{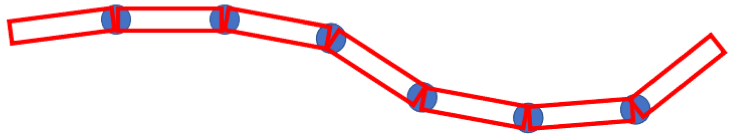}
    \caption{The representation of a DOO as a chain of fixed-length cylinders connected by spherical joints.}
    \label{fig:doo-representation}
\end{figure}

A commonly-used approach for representing DOOs is to model them as a chain of fixed-length cylindrical segments connected by spherical joints. This simple model can easily integrate with an efficient dynamics model to simulate or predict the object's behavior. While the chosen model does not affect the ideas described in the proposed method, our method utilizes this model. In our application, the length of each segment is represented by $l_s$, and there is no gap between the segments (i.e., each segment starts precisely where the neighbor segment ends).
Figure~\ref{fig:doo-representation} illustrates the fixed-length cylinder chain model used in this work.

The focus of this work is to provide the cylinder chain representation of a deformable one-dimensional object seen in the camera frame (which can be RGB or RGB-D/3-D). The output chain model should predict and fill the path taken by the DOO under the occlusions and return a single chain object. Figure~\ref{fig:results-1} shows this objective.
\section{Proposed Method} \label{sec:method}

\begin{figure}[t]
    \centering
    \includegraphics[width=0.9\linewidth]{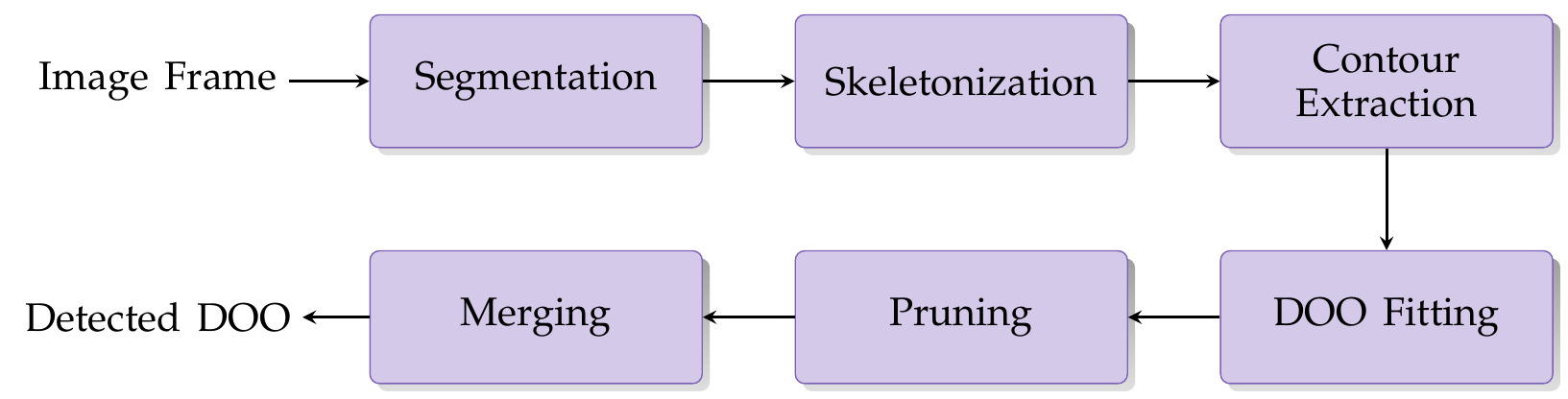}
    \caption{The high-level overview of the proposed method for detection of deformable one-dimensional objects.}
    \label{fig:overview}
\end{figure}

The DOO detection method in this work takes the camera frame and performs a sequence of processing steps to output a single object in chain representation (described in Section~\ref{sec:problem}). Figure~\ref{fig:overview} shows a high-level overview of the method.

The first three steps in the proposed method are well-known processes. Our algorithm can work with different segmentation, skeletonization, and contour extraction approaches. The choice depends on the task at hand. The last three processing steps are the contributions of our algorithm. 

Algorithm~\ref{alg:doo-detection} shows the pseudo-code of our approach. This section describes each of those steps in more detail.

\begin{algorithm}[!t]
\caption{Deformable one-dimensional object detection.}
\label{alg:doo-detection}
\begin{algorithmic}[1]
\LineComment {Detects and extract DOOs from the input image frame.}
\Function{DetectDOO}{frame}
    \LineComment{Segment the image to extract the DOO region}
    \State $segmented\_img \gets \textsc{Segment}(frame)$
    \LineComment{Extract the skeletons of the segmented regions}
    \State $thinned\_img \gets \textsc{Skeletonize}(segmented\_img)$
    \LineComment{Extract the contours from the skeletons}
    \State $contours \gets \textsc{ExtractContours}(thinned\_img)$
    \LineComment {Fit DOO chains to all contours}
    \State $Chains \gets \emptyset$
    \ForAll {$c \in contours$}
        \State $fitted\_chains \gets \textsc{FitDOO}(c)$
        \State $Chains$.insert$(fitted\_chains)$
    \EndFor
    \LineComment {Prune all the overlapping segments}
    \State $Chains \gets \textsc{Prune}(Chains)$
    \LineComment {Merge the DOO chains into a single DOO}
    \While {$Chains.length > 1$}
        \State $C_1, C_2 \gets \textsc{FindBestMergeMatch}(Chains)$
        \State $C_{merged} \gets \textsc{MergeChains}(C_1, C_2)$
        \State $Chains$.remove$(\{C_1, C_2\})$
        \State $Chains$.insert$(C_{merged})$
    \EndWhile
    \State \Return $Chains[0]$ \NewComment {Return the final DOO chain}
\EndFunction
\end{algorithmic}
\end{algorithm}

\subsection{Segmentation}

A vital step in extracting the complete DOO from the input camera image is segmentation. This step aims to filter the image data to extract the DOO portions and exclude all other data. 
%
%

The simplest segmentation methods include color-based filtering and background subtraction, which can work in lab settings, but more complex methods are required for real-world applications.

The medical research community provides an extensive set of segmentation methods to address vein and vessel detection, which can be used here directly or with small modifications. Such methods include both model-based~\cite{BFb0056195, KRISSIAN2000130, 28493, Wang2021, 10605, Noble2011} and learning-based~\cite{72a4e1c53c, 32226} approaches and are often robust to clutter in the input image and can work in low signal-to-noise conditions.

\begin{figure}[!t]
\centering
    \begin{subfigure}[b]{0.23\textwidth}
        \includegraphics[width=\textwidth]{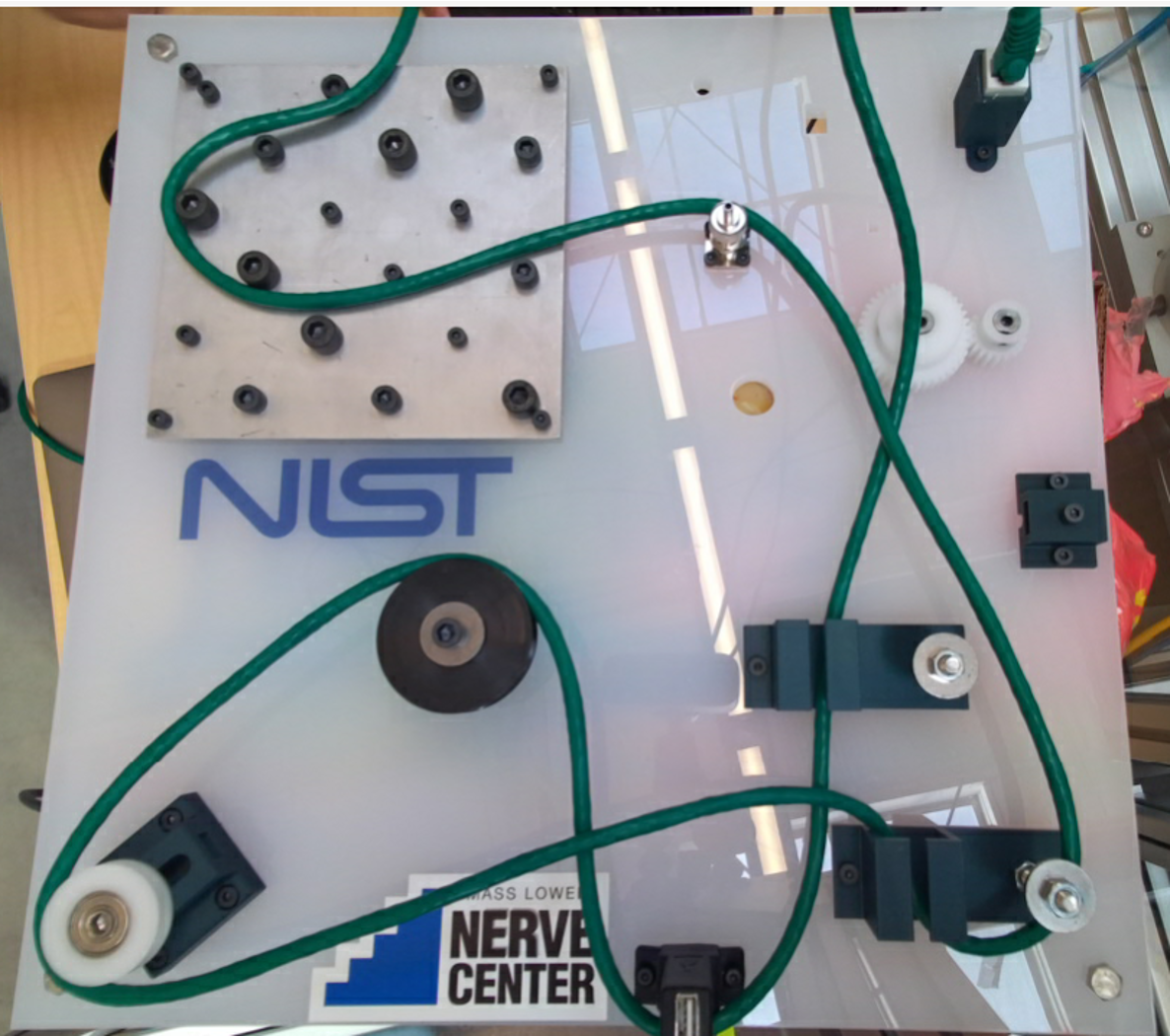}
        \caption{~}
        \label{fig:original-board}
    \end{subfigure}
    \hfill
    \begin{subfigure}[b]{0.23\textwidth}
        \includegraphics[width=\textwidth]{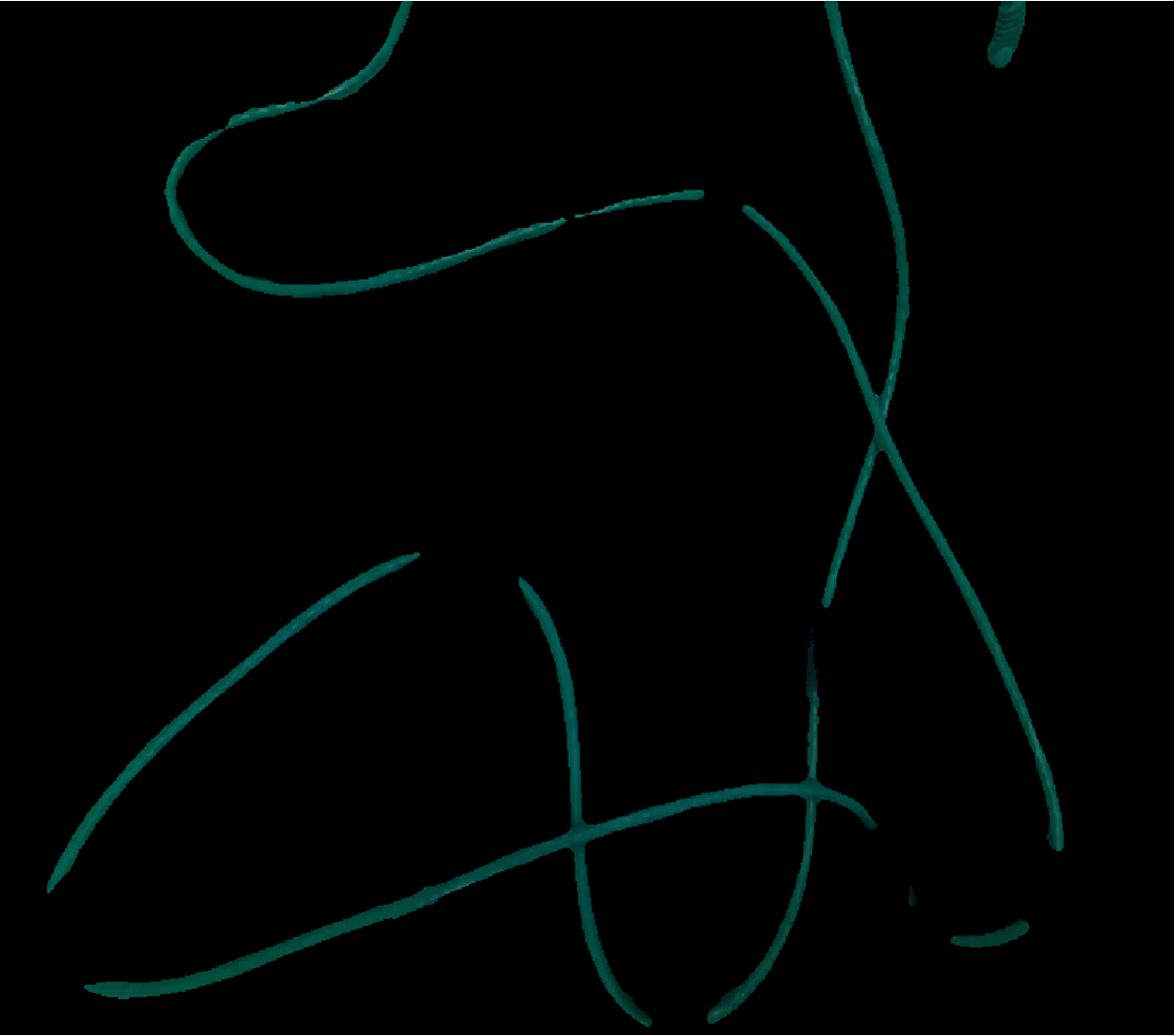}
        \caption{~}
        \label{fig:segmented-board}
    \end{subfigure}
    \caption{The segmentation of a DOO in a camera frame. (a) The original image. (b) The segmentation result.}
    \label{fig:segmentation}
\end{figure}

The requirement for the segmentation method for our work is to filter the DOO data conservatively, i.e., ideally, it should eliminate all the unrelated data even if it removes some of the DOO data. Figure~\ref{fig:segmentation} illustrates the segmentation of an example cable in the camera frame.

\subsection{Topological Skeletonization}
    
Skeletonization transforms each segmented connected component into a set of connected pixels with single-pixel width called a \textit{skeleton}. It is commonly used in the pre-processing stage of various applications ranging from Optical Character Recognition (OCR) to human motion tracking, fingerprint analysis, and various medical imaging analysis~\cite{SAHA20173, keipour2013omnifont}.

\begin{figure}[t]
\centering
    \begin{subfigure}[b]{0.21\textwidth}
        \includegraphics[width=\textwidth, height=3cm]{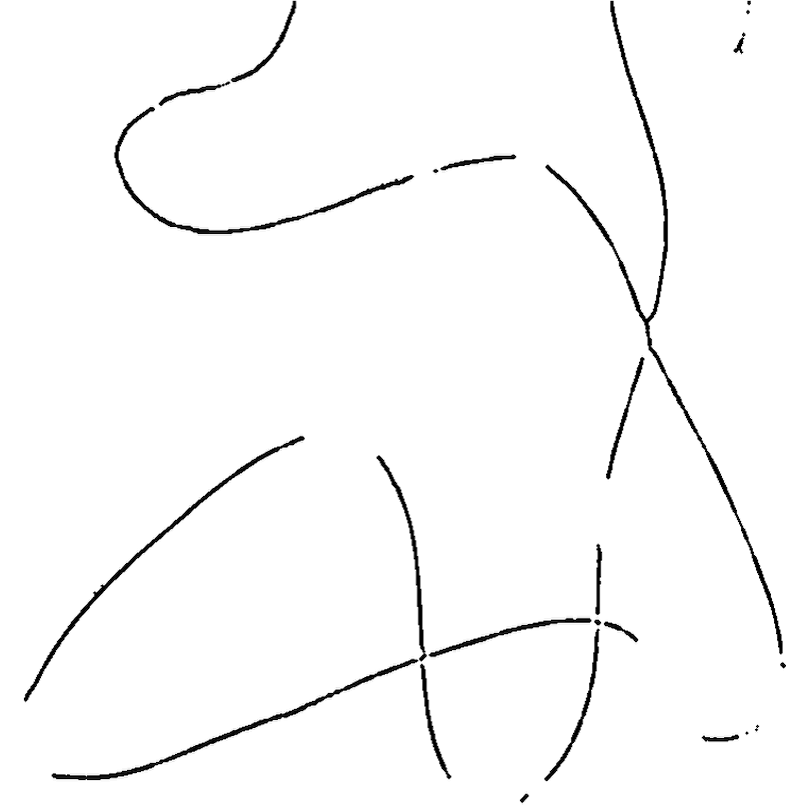}
        \caption{~}
        \label{fig:skeletonization}
    \end{subfigure}
    \hfill
    \begin{subfigure}[b]{0.21\textwidth}
        \includegraphics[width=\textwidth, height=3cm]{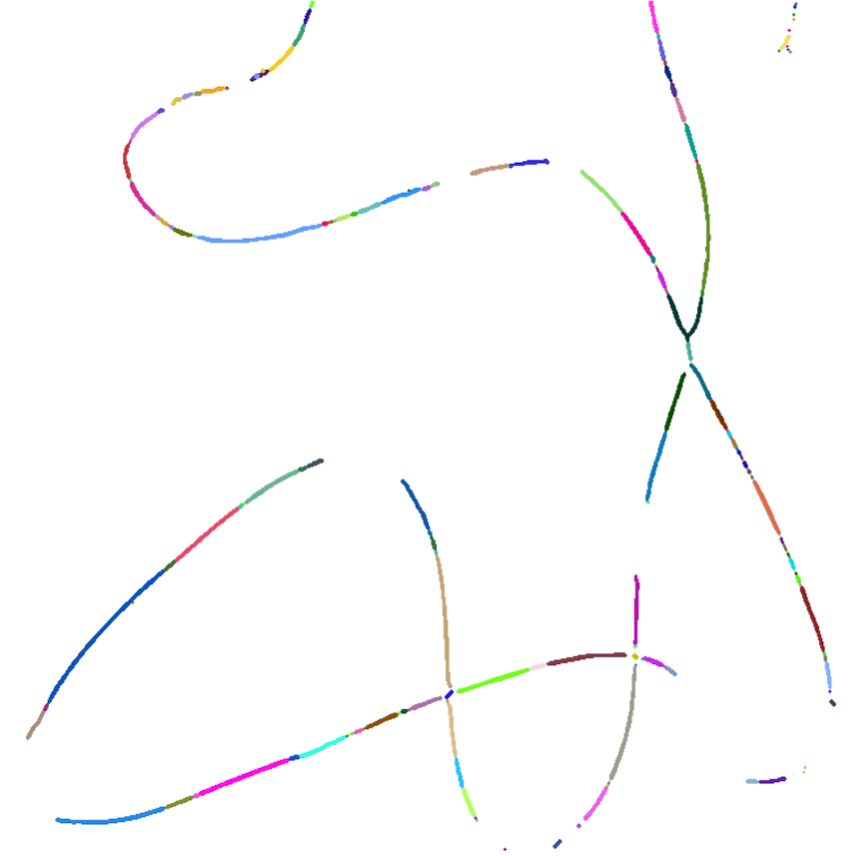}
        \caption{~}
        \label{fig:contour-extraction}
    \end{subfigure}
\label{fig:skeletonization-and-contours}
\caption{(a) The skeletonization of a segmented deformable one-dimensional object. (b) The contours extracted from the skeleton. Each contour is drawn with a different color.}
\end{figure}

Our algorithm has two requirements for choosing the skeletonization method: a) the skeleton of a connected component should remain a connected component; b) only one branch should be returned per actual branch (i.e., multi-branching of a single skeleton branch should be avoided). The skeletonization algorithm chosen for this step should inherently respect the two constraints. Figure~\ref{fig:skeletonization} shows the skeletonization of the segmented example of Figure~\ref{fig:segmentation}.
    
\subsection{Contour Extraction}

A contour (a.k.a., boundary) is an ordered sequence of the pixels around a shape. Extracting contours from an image is utilized in many applications ranging from shape analysis to semantic segmentation and image classification~\cite{Gong2018, 9484730}. 

\begin{figure}[!t]
    \centering
    \includegraphics[width=0.15\textwidth]{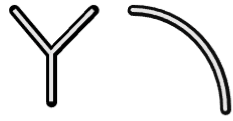}
    \caption{Contour types extracted from a skeletonized image. The black area around the gray skeleton is the contour.}
    \label{fig:contour-types}
\end{figure}

Ideally, the contour extraction method applied to the skeletons should result in one contour per skeleton branch. However, in practice, the contour extraction methods can result in several contours per branch and some contours containing multiple branches. Moreover, a contour contains the closed boundary \textit{around} the skeleton and not the actual skeleton pixels. Figure~\ref{fig:contour-types} shows different types of contours that can be extracted from a skeleton piece.

Our DOO detection method can handle the above-mentioned common issues raised from the contour extraction methods. Therefore, many of the existing contour extraction methods can be used with our algorithm regardless of their output limitations. Figure~\ref{fig:contour-extraction} presents the result of contour extraction.

Contours facilitate traversing points along the skeleton and simplify understanding the connections in the branches. If an ordered set of pixels for each skeleton branch is obtained from the skeletonization method or other means, the contour extraction step can be skipped.


\subsection{Fitting DOO Segments} \label{sec:segment-fitting}

The next step is to fit a chain of fixed-length segments to each contour. A contour can be a single branch, or it may contain multiple branches (see Figure~\ref{fig:contour-extraction}). The pixel sequence for a contour returned by a typical contour extraction method starts from one of the tips and ends with a sharp turn back at the start point.

Let us call the latest added segment as $s$, the current segment as $s'$, the first point in the contour sequence as the starting point $p_s$ of $s'$, and the point currently being traversed as $p_c$. Let us also define $\Vec{s}$ as the vector in the direction of segment $s$, starting at its start point and pointing towards its end point. Therefore, starting with an empty DOO chain, the points are traversed while the distance $\|p_c - p_s\|_2$ is less than $l_s$. Then a new segment of length $l_s$ is added to the DOO chain from point $p_s$ in the direction of $p_c$, the new segment's end-point $p_e$ is saved as the next segment's start point $p_s$, and the traversal continues.

Three conditions may happen during the traversal:

\begin{enumerate}[leftmargin=*]

    \item Vector $\Vec{s'}$ is close to vector $\Vec{s}$: The new segment is added to the current DOO chain in this case.
    
    \item Vector $\Vec{s}'$ is close to vector $-\Vec{s}$: It means that the traversal has gone over a branch end. In this case, the current DOO is recorded without the new segment, and the new segment is discarded. The traversal continues from the branch tip with a new empty DOO chain.

    \item The last point in the contour sequence is reached: it means that the traverse has returned to the start point, and the traverse can be terminated. There may be cases where the whole contour length is less than the segment size. In these cases, no new DOO chains will be generated.
\end{enumerate}

Algorithm~\ref{alg:contour-traversal} shows the pseudo-code for the described steps.

\begin{algorithm}[!t]
\caption{Traversing contours for DOO chain creation.}
\label{alg:contour-traversal}
\begin{algorithmic}[1]
\LineComment {This function traverses a contour and returns all DOO chains created from the contour.}
\Function{TraverseContour}{\textit{contour}}
    \LineComment {Initialize the collection of DOO chains}
    \State $Collection \gets \emptyset$
    \LineComment {Initialize the start point and the next DOO chain}
    \State $p_s \gets contours[0]$\quad,\quad$chain \gets \emptyset$
    \LineComment {Initialize the tip point}
    \State $p_t \gets p_s$\quad,\quad$update\_tip \gets True$
    \State
    \LineComment {Traverse over all the points in the contour}
    \ForAll {$p_c \in contour$}
        \LineComment {Create a new segment if $p_c$ is far enough}
        \If {$\textsc{Dist}(p_c, p_s) \geq l_s$}
            \State $p_e \gets p_s + l_s \times (p_c - p_s) / \|p_c - p_s\|$
            \State $s' \gets \textsc{CreateSegment}(p_s, p_e)$
            \If {$chain = \emptyset$ or $\textsc{Angle}(s, s')$ is small}
                \LineComment{Add the segment if it is the first in chain}
                \LineComment {or if the direction has not changed much}
                \State $chain.$Insert$(s')$
                \State $s \gets s'$\quad,\quad$p_s \gets p_e$\quad,\quad$p_t \gets p_e$
            \Else
                \LineComment{Start a new chain if direction has changed}
                \State $Collection.$Insert$(chain)$
                \State $chain \gets \emptyset$\quad,\quad$p_s \gets p_t$
            \EndIf
            \LineComment{Start updating tip point}
            \State $update\_tip \gets True$
        \ElsIf {$\textsc{Dist}(p_c, p_s) \geq \textsc{Dist}(p_t, p_s)$}
            \LineComment{Update the tip point}
            \IIf {$update\_tip = True$} {$p_t \gets p_c$} \EndIIf
        \Else
            \LineComment{Stop updating tip point}
            \State $update\_tip \gets False$
        \EndIf
    \EndFor
    \LineComment {Add the last generated chain}
    \State $Collection.$Insert$(chain)$
    \State \Return $Collection$ \NewComment {Return all the chains at the end}
\EndFunction
\end{algorithmic}
\end{algorithm}


\subsection{Pruning}

The segment fitting algorithm of Section~\ref{sec:segment-fitting} returns multiple overlapping DOO chains for each part of the object. It is desired to prune the overlapping segments to reduce the total number of segments and simplify the further steps by assuming that no two segments overlap.

\begin{figure}[!t]
\centering
    \begin{subfigure}[b]{0.23\textwidth}
        \includegraphics[width=\textwidth]{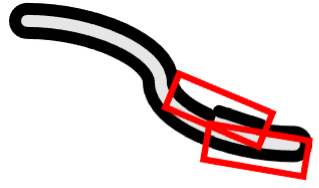}
        \caption{~}
        \label{fig:segment-overlap}
    \end{subfigure}
    \hfill
    \begin{subfigure}[b]{0.22\textwidth}
        \includegraphics[width=\textwidth]{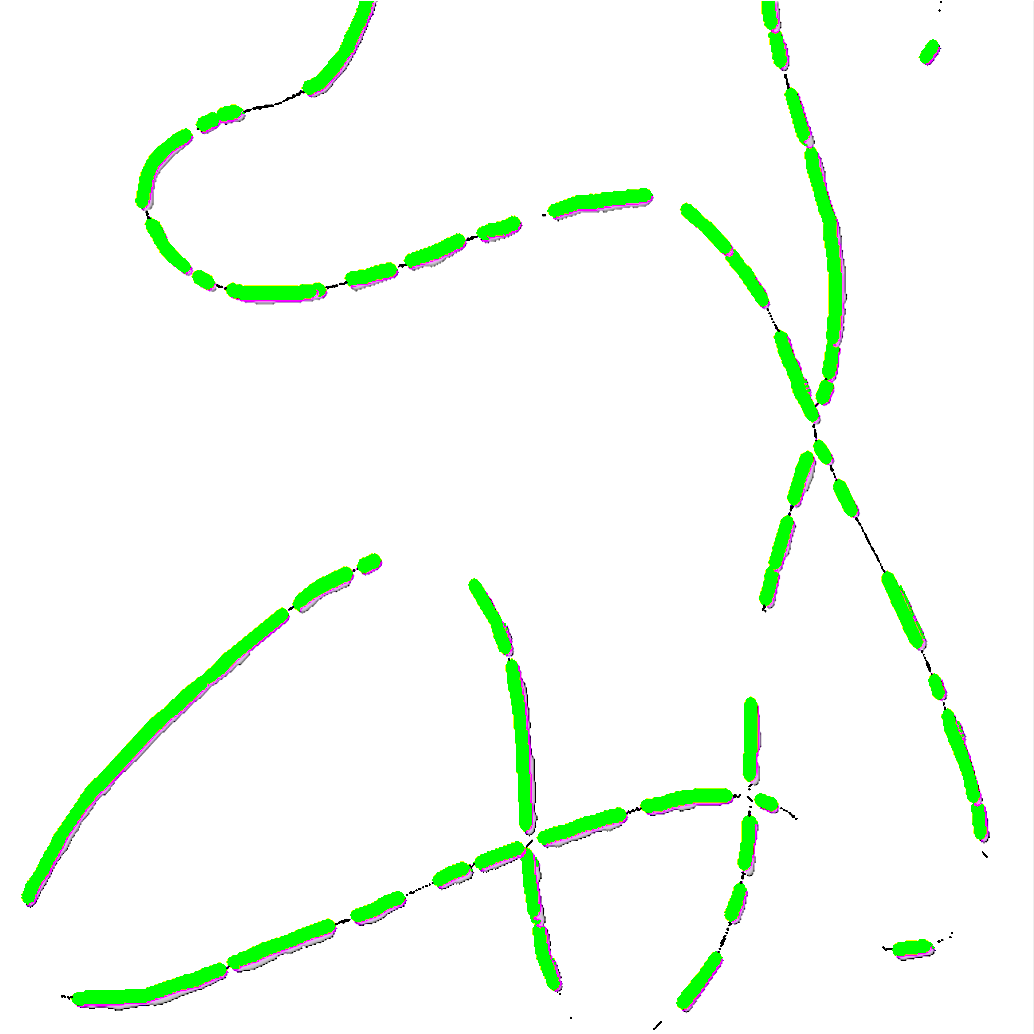}
        \caption{~}
        \label{fig:pruning}
    \end{subfigure}
    \caption{(a) An illustration of overlapping segments around a skeletonized deformable one-dimensional object. (b) The result of segment fitting from contours and pruning for a deformable one-dimensional object of Figure~\ref{fig:skeletonization}.}
    \label{fig:fitting-prining}
\end{figure}

To define the overlap of two segments, each segment can be assumed as a rotated rectangle in the 2-D case and a square cuboid (a cuboid with two square faces) for the 3-D case. The length of the rectangle and cuboid is the segment length $l_s$, and the rectangle's width is 3 pixels or higher. The reasoning for the choice of the width is that the width of the skeleton is generally 1 pixel with occasional width of 2 pixels (depends on the choice of the thinning method). The contour is the boundary around the skeleton, and for the two rectangles on the two sides of the skeleton to overlap, they need to be at least 3 pixels wide. In practice, any small number greater than 3 pixels should work well for pruning the overlapping segments. For the 3-D case, the width of the cuboid can be chosen similarly, i.e., it should be at least the total of the width of the contour layer and the maximum width of the skeleton. Once the segments are defined, the geometric intersection of two rotated rectangles or cuboids is used to find the overlap. Figure~\ref{fig:segment-overlap} shows how two segments can overlap for the same skeleton.

A heuristic that has shown performance improvements in the subsequent stages removes the segment from the shorter chain when two segments overlap. This heuristic will result in many chains being quickly emptied, which reduces the overall number of DOO chains in the collection.

Figure~\ref{fig:pruning} shows the chains resulting from segment fitting and then pruning of the contours in Figure~\ref{fig:contour-extraction}.


\subsection{Merging}

Once we have a collection of DOO chains, they should be merged to fill the gaps and form a single object. A gap can result from an occlusion or an imperfect segmentation and should be filled in a way that follows the natural curve of the deformable object. 

The merging process can be performed iteratively, connecting two chains at a time until all the chains are merged into a single deformable one-dimensional object. Each iteration can be broken down into two steps:

\begin{enumerate}
    \item Choose the best two chains for merging
    \item Connect the selected chains
\end{enumerate}

The following subsections describe choosing the best chains and properly connecting them with their natural bend.


\subsection{Merging: Choosing the Best Matches} \label{sec:method:merging:choosing}

To choose the best chains to connect, we define a new cost function $C_M(\cdot)$ that calculates the cost of connecting any two chain ends. Considering that there are two ends for each chain, there will be four cost values for connecting the two ends for any two chains. The lowest among the four values is the cost of connecting the two chains. 

Given an end segment $s_1$ of the first chain and an end segment $s_2$ of the second chain, three separate partial costs are defined and then combined to create the total cost function $C_M(\cdot)$:

\begin{itemize}[leftmargin=*]
    \item Euclidean Cost $C_E$: This measure incurs costs to two chain ends based on their Euclidean distance to deter the early connection of far away ends (Figure~\ref{fig:merge-costs-e}). Having the end segments $s_1$ and $s_2$, this cost can be calculated as:
    
    \begin{equation}
        C_E\left(s_1, s_2\right) = \|s_1.end - s_2.end\|_2,
    \label{eq:cost-euclidean}
    \end{equation}
    \noindent where $\|\cdot\|_2$ is the norm of the resulting vector and $s.end$ is the end point of the segment (end point of the chain).
    
    \begin{figure}[!t]
    \centering
        \begin{subfigure}[b]{0.115\textwidth}
            \includegraphics[width=\textwidth]{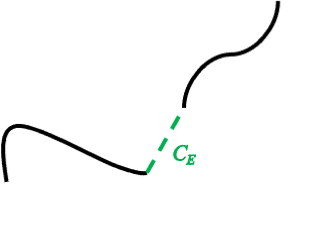}
            \caption{~}
            \label{fig:merge-costs-e}
        \end{subfigure}
        \hfill
        \begin{subfigure}[b]{0.115\textwidth}
            \includegraphics[width=\textwidth]{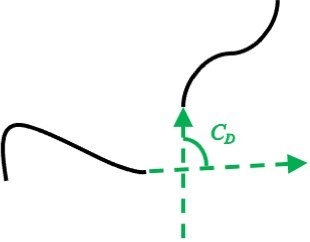}
            \caption{~}
            \label{fig:merge-costs-d}
        \end{subfigure}
        \hfill
        \begin{subfigure}[b]{0.23\textwidth}
            \includegraphics[width=\textwidth]{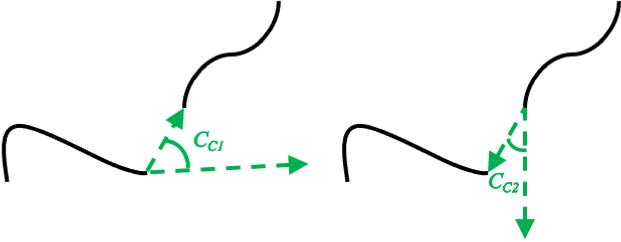}
            \caption{~}
            \label{fig:merge-costs-c}
        \end{subfigure}
    \caption{Illustration of different partial costs of merging two chain ends. (a) Euclidean cost. (b) Direction cost. (c) Curvature costs from first chain to the second and from the second chain to the first.}
    \label{fig:merge-costs}
    \end{figure}

    \item Direction Cost $C_D$: This measure incurs costs to two chain ends based on the difference of the direction of the first with the opposite direction of the second (Figure~\ref{fig:merge-costs-d}). This cost discourages the connection of chains that are not facing each other. Having the end segments $s_1$ and $s_2$, this cost can be calculated as:
    
    \begin{equation}
        C_D\left(s_1, s_2\right) = \left|\arccos{\left(\frac{-\Vec{s_1}\cdot \Vec{s_2}}{\|s_1\|\|s_2\|}\right)}\right|,
    \label{eq:cost-direction}
    \end{equation}
    
    \noindent where $|\cdot|$ is the absolute value, $\|\cdot\|$ is the norm of the vector (size of the segment), $\Vec{s}$ is the vector along the segment $s$ starting from the start of the segment and ending at the end of the segment (i.e., the end of the chain), and $\cdot$ is the inner product operator.

    \item Curvature Cost $C_C$: This measure incurs costs to two chain ends based on how much curvature is needed to connect them. It is calculated as the higher cost between bending the first chain's end segment towards the second chain's end segment and vice versa (Figure~\ref{fig:merge-costs-c}). The measure is defined to discourage connections requiring excessive bending and to encourage smooth connections. Having the end segments $s_1$ and $s_2$, this cost can be calculated as:
    
    \begin{equation}
        \begin{split}
            &C_{C1}\left(s_1, s_2\right) = \left|\arccos{\left(\frac{\Vec{s_1}\cdot \Vec{s_{21}}}{\|s_1\|\|s_{21}\|}\right)}\right| \\
            &C_{C2}\left(s_1, s_2\right) = \left|\arccos{\left(\frac{\Vec{s_2}\cdot \Vec{s_{12}}}{\|s_2\|\|s_{12}\|}\right)}\right| \\
            &C_C\left(s_1, s_2\right) = \max{(C_{C1}, C_{C2})},
        \end{split}
        \label{eq:cost-curvature}
    \end{equation}
    
    \noindent where $s_{nm}$ is a shorthand for $s_n.end - s_m.end$.
\end{itemize}

Having the three cost values for the ends of two chains, the total cost of these ends is computed as:

\begin{equation}
    \begin{split}
        C_M(&s_1, s_2) = \\
        &\mathscr{F}\Big(C_E\left(s_1, s_2\right), C_D\left(s_1, s_2\right), C_C\left(s_1, s_2\right)\Big),
    \end{split}
    \label{eq:cost-merge}
\end{equation}

\noindent where $\mathscr{F}$ is the function combining the three values. In practice, we learned that the weighted sum of the values works well, and even after manually choosing a simple weight set, the algorithm works for almost all kinds of situations (see Section~\ref{sec:tests} for our test values). With the weighted sum, the Equation~\ref{eq:cost-merge} reduces to:

\begin{equation}
    \begin{split}
        &C_M(s_1, s_2) = \\
        &w_e \cdot C_E\left(s_1, s_2\right) + w_d \cdot C_D\left(s_1, s_2\right) + w_c \cdot C_C\left(s_1, s_2\right)
    \end{split}
    \label{eq:cost-merge-weighted-sum}
\end{equation}

After calculating the four costs of all end combinations of the two chains, the minimum of those costs is the cost of merging the two DOO chains. Once the costs for all pairs of chains are calculated, the two chains with the lowest total merging cost are chosen for merging. 

Note that the choice for the cost function of Equation~\ref{eq:cost-merge} is to encourage the connection of closer chains that align well and can connect smoothly. Choosing a single measure such as minimum curvature would result in unwanted connections of farther chains that align perfectly over closer chains that are slightly misaligned.


\subsection{Merging: Connecting Two Chains}

Once two chains $C_1$ and $C_2$ are selected for connection (see Section~\ref{sec:method:merging:choosing}), the gap between the two chains should be filled with a new chain $C_{new}$ in a way that it follows the expected curve of the deformable object. Our experiments show that any deformable object can take almost any curve given different pressure points, forces, tensions, and the object's condition. However, it is possible to have an educated \textit{guess} on how the object behaves. For this purpose, we calculate the "natural" curvature required for the new chain $C_{new}$, which connects the desired end of $C_1$ to the desired end of $C_2$.

\begin{figure}[!t]
\centering
    \begin{subfigure}[b]{0.15\textwidth}
        \includegraphics[width=\textwidth]{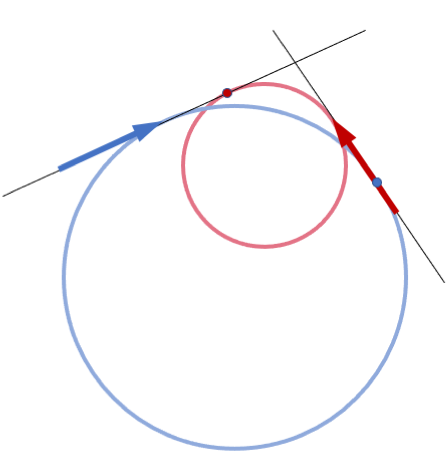}
        \caption{~}
        \label{fig:merge-cases-1}
    \end{subfigure}
    \hfill
    \begin{subfigure}[b]{0.15\textwidth}
        \includegraphics[width=\textwidth]{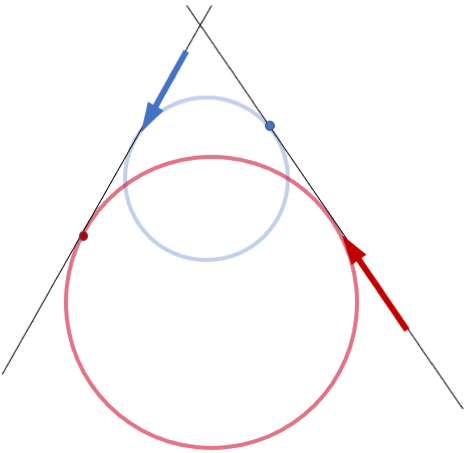}
        \caption{~}
        \label{fig:merge-cases-2}
    \end{subfigure}
    \hfill
    \begin{subfigure}[b]{0.15\textwidth}
        \includegraphics[width=\textwidth]{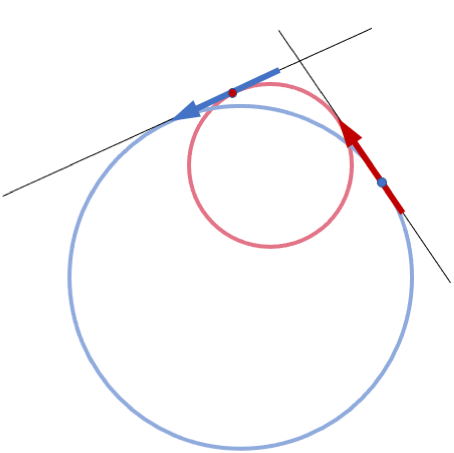}
        \caption{~}
        \label{fig:merge-cases-0}
    \end{subfigure}
\caption{Illustration of different merging scenarios with the two circles tangent to the line passing the end points of the two chains, each circle passing through one of the end points. Arrow ends and directions represent the end points and end directions of the chains. (a) Exactly one of the circles passing through the end point of a chain is touching the other line ahead of the other chain. (b) Both the circles passing through the end points of the chains are touching the other line ahead of the other chain. (c) None of the circles passing through the end point of a chain are touching the other line ahead of the other chain.}
\label{fig:merge-cases}
\end{figure}

To compute the "natural" curvature, we assume that the new chain $C_{new}$ starts in the same direction as the two desired chain ends. In other words, at each end, $C_{new}$ initially follows the direction of the last segment of the chain to which it is connected. On the other hand, we assume that when it is possible, the deformable one-dimensional object will follow a curve with a constant turn rate (i.e., constant radius). With these assumptions, we can find two circles tangent to the lines passing through the two chain ends, each passing through one of the chain end-points. Based on triangle similarity theorems, the radii of the two circles are proportional to the distances of the chain ends from the intersection point. Figure~\ref{fig:merge-cases} illustrates this idea.

Let us define the end-points we desire to connect on chains $C_1$ and $C_2$ as $e_1$ and $e_2$, respectively. We call the lines passing through $e_1$ and $e_2$ in the direction of $C_1$ and $C_2$ ends as $l_1$ and $l_2$, respectively. Finally, we define the circles passing through $e_1$ and $e_2$ as $c_1$ and $c_2$, and the points they touch on the other line as $t_1$ and $t_2$, respectively. Note that points $e_1$ and $t_2$ will be lying on line $l_1$, while points $e_2$ and $t_1$ are on line $l_2$.

Without loss of generality, let us assume that in Figure~\ref{fig:merge-cases}, circle $c_1$ is the red circle, the blue circle is $c_2$, the red dot is $t_1$, the blue dot is $t_2$, the red arrow's end point (arrow side) is $e_1$, the blue arrow's end point is $e_2$, the line passing $e_1$ is $l_1$ and the line passing $e_2$ is $l_2$.

It can be proven that the distance between $e_2$ and $t_1$ is equal to the distance between $e_1$ and $t_2$. However, each $t_1$ and $t_2$ can be lying on lines $l_2$ and $l_1$ ahead or behind $e_2$ and $e_1$, creating three different situations:

\begin{itemize}[leftmargin=*]
    \item Either $t_1$ is ahead of $e_2$ or $t_2$ is ahead of $e_1$, but not both (Figure~\ref{fig:merge-cases-1}). Not surprisingly, a majority of connections in a typical application would be of this type.
    In this case the blue circle $c_2$ that touches $l_1$ at point $t_2$ behind $e_1$ is discarded and we use the radius of the red circle $c_1$ for the turn radius of the new chain $C_{new}$. This chain will be composed of the arc of $c_1$ from $e_1$ to $t_1$ and the line from $t_1$ to $e_2$. Figure~\ref{fig:merge-fill-1} shows this scenario's solution.

    \begin{figure}[!t]
    \centering
        \begin{subfigure}[b]{0.11\textwidth}
            \includegraphics[width=\textwidth]{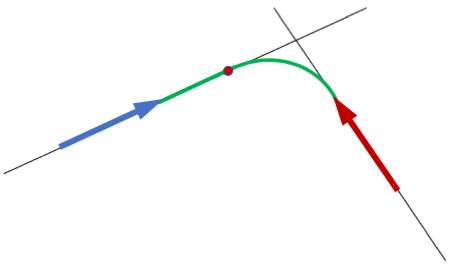}
            \caption{~}
            \label{fig:merge-fill-1}
        \end{subfigure}
        \hfill
        \begin{subfigure}[b]{0.11\textwidth}
            \includegraphics[width=\textwidth]{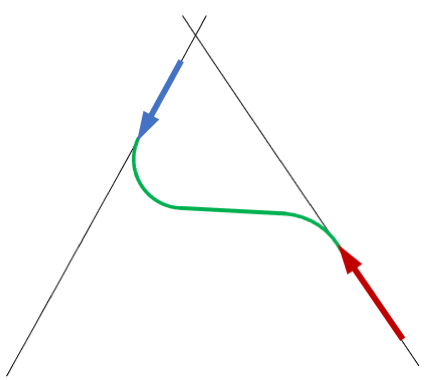}
            \caption{~}
            \label{fig:merge-fill-2}
        \end{subfigure}
        \hfill
        \begin{subfigure}[b]{0.22\textwidth}
            \includegraphics[width=\textwidth]{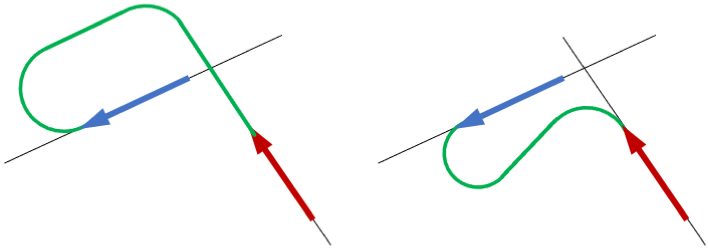}
            \caption{~}
            \label{fig:merge-fill-0}
        \end{subfigure}
    \caption{Suggested solutions for different merging cases illustrated in Figure~\ref{fig:merge-cases}.}
    \label{fig:merge-fill}
    \end{figure}
    
    \item Both $t_1$ and $t_2$ are ahead of $e_2$ and $e_1$ (Figure~\ref{fig:merge-cases-1}). 
    In this case, the new chain $C_{new}$ is composed of an arc on each end ($e_1$ and $e_2$) and a line tangent to the arcs. The turn radius is as desired or can be experimentally determined for the DOO, and it should be large enough to allow the "natural-looking turn." However, the radius should be small enough so that the direction of the line tangent to the two arcs is close to the direction of the line connecting $e_1$ to $e_2$. Finally, we suggest the same turn radius for both ends. Figure~\ref{fig:merge-fill-2} shows this scenario's solution.

    \item Both $t_1$ and $t_2$ are behind of $e_2$ and $e_1$ (Figure~\ref{fig:merge-cases-0}). This case has two suggested solutions that depend on the conditions.
    In both solutions, similar to the previous case, the new chain $C_{new}$ is composed of an arc on each end ($e_1$ and $e_2$) and a line tangent to the arcs. The turn radius is as desired or can be experimentally determined for the DOO. However, depending on external conditions, $C_{new}$ can fill the gap from outside or inside the region between the two chains. Figure~\ref{fig:merge-fill-0} shows this scenario's solutions.

\end{itemize}

Note that, in all scenarios, there can be infinite correct solutions, which depend on the conditions. In practice, the suggested solutions result in good fits with the ground truth and can be used in most conditions without any modification.

\begin{figure}[t]
    \centering
    \includegraphics[width=0.5\linewidth]{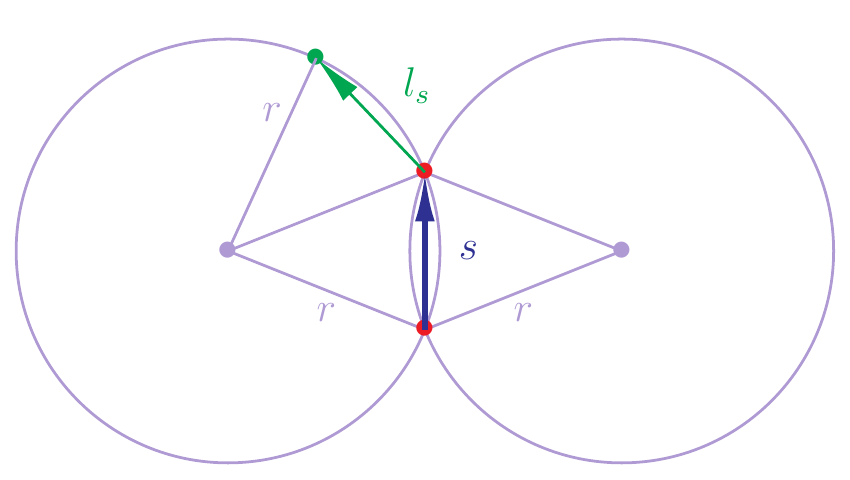}
    \caption{Adding a new DOO segment with length $l_s$ with a desired constant turn radius of $r$ to the end of segment $s$.}
    \label{fig:enforce-turn}
\end{figure}

Once the new chain $C_{new}$ is obtained to fill the gap between the two chain ends $C_1$ and $C_2$, it can be added to the ends of the chains to connect them. $C_{new}$ is a combination of constant-radius arcs and lines. Adding a DOO segment for the line sections is trivial and will not be explained. To add a segment that follows the desired turn, we use the last segment $s$ on the chain. Knowing the start and end-points of this segment $s$, we can calculate two circles with the desired radius that pass through these points. Knowing the direction of the turn, one circle is eliminated, and the new point on the remaining circle at the segment distance $l_s$ of the segment's end-point is used to create the new segment $s'$ that is added to the end of the chain. Figure~\ref{fig:enforce-turn} shows how the new segment can be added with the desired turn radius.

\subsection{Notes on the Proposed Method}

The merging process continues connecting the chains, two at a time, until all the chains are merged into one single chain, which is the detected DOO represented by the chain of fixed-length cylindrical segments connected by passive spherical joints. This representation can be used as the input to routing and manipulation systems for the desired application. 

Each segmented image will be processed into a single DOO. When there are multiple DOOs present in the camera frame, the easiest way to detect them as separate DOOs is to have separate segmentation for them. For example, if there are multiple cables with different colors in the image, a color-based segmentation can have two segmented images. In case there are multiple DOOs that cannot be segmented separately, the proposed algorithm can still help process them into separate DOO outputs. Note that the algorithm is greedy in the sense that it first goes for the best-matched chains. With multiple DOOs, there is a high chance that the chains related to separate DOOs do not give a good fit. As a result, for example, when there are only two chains left, there is a high chance that the two chains are the two separate DOOs. Therefore, it is enough to stop the merging process when the desired number of chains is left.

The proposed method is general and can be used with both 2-D and 3-D image data to provide the deformable object's representation in 2-D or 3-D.

Finally, the final output of the proposed method is a single chain of segments that does not keep track of the parts seen in the frame vs. the occluded parts. There are two ways to mark the parts of the chain that are related to the occlusions: 

\begin{enumerate}[leftmargin=*]
    \item Map the segmented parts on the final detection to determine the occluded parts of the DOO chain.
    
    \item During the process, when merging two chains, if the gap is equal or longer than the segment length $l_s$, the newly added chain $C_{new}$ is marked as occluded. The reason for skipping smaller gaps is that many gaps shorter than $l_s$ are created during the pruning process. 
\end{enumerate}

Both approaches ultimately depend on the accuracy of the segmentation. The first approach is simple but may mislabel an occluded part as visible in a multilayer setup. On the other hand, the second approach tends to be more accurate in multilayer settings but may skip more minor occlusions.

\section{Experiments and Results} \label{sec:tests}

The proposed was implemented for 2-D images in Python 3. We used the color-based segmentation of the DOO region. This approach generally tends to include extra areas around the DOO and other regions with similar color hues to the DOO. We chose conservative thresholds to exclude any non-DOO regions. This results in some DOO data being excluded; However, our experiments have shown the DOO detection to have challenges when extra regions are included but to work when some data is lost in segmentation. The same principle is advised for other segmentation methods choices, and those methods' parameters should be chosen conservatively to remove the irrelevant regions.

We used a well-known morphological thinning method for skeletonization~\cite{1164959}. The algorithm proposed by Suzuki and Abe~\cite{suzuki1985topological} and provided in the OpenCV library is used to extract contours. All our tests use $w_e = 1$, $w_d = 100$ and $w_c = 100$ values for the cost function of Equation~\ref{eq:cost-merge-weighted-sum} and the segment length $l_s$ is chosen as 10 pixels. The weights are chosen manually to focus on shorter distances while heavily discouraging non-matching segment directions and excessive bending. Different weights result in some types of incorrect connections increasing while the number of other types decreases. A better set can be found using a more methodical approach and optimization for the desired applications. Similarly, the segment length was not chosen optimally. In general, a shorter segment length can capture the contour ends better and create segments from smaller contours, leading to better curves in filling gaps; however, it increases the number of segments and the average number of incorrect connections. On the other hand, a longer segment length has the potential of not following the curves well and ignoring small contours. However, it tends to reduce the number of incorrect connections and improve the detection speed by reducing the number of segments.

\begin{figure}[!t]
\centering
    \begin{subfigure}[b]{0.48\textwidth}
        \includegraphics[width=0.32\textwidth, height=3cm]{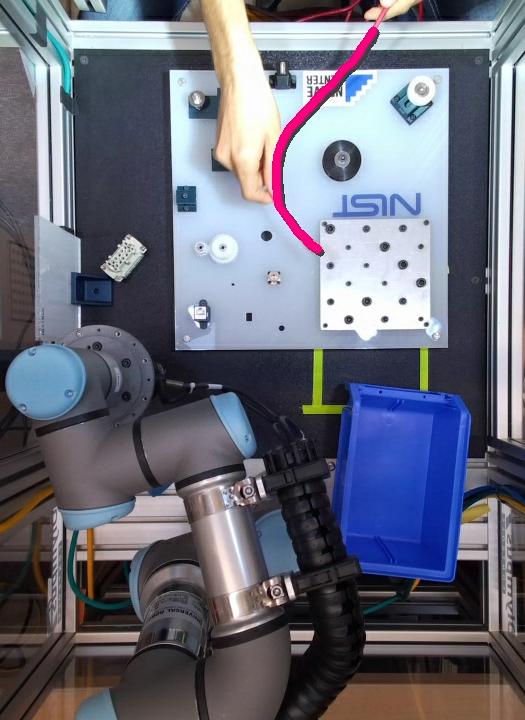}
        \hfill
        \includegraphics[width=0.32\textwidth, height=3cm]{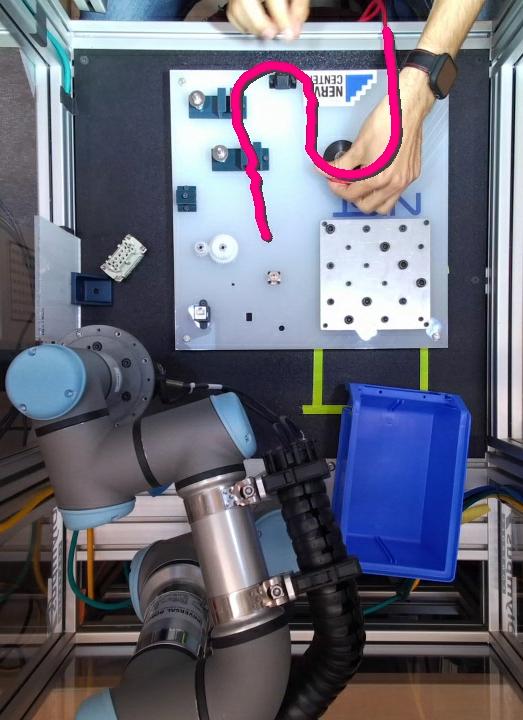}
        \hfill
        \includegraphics[width=0.32\textwidth, height=3cm]{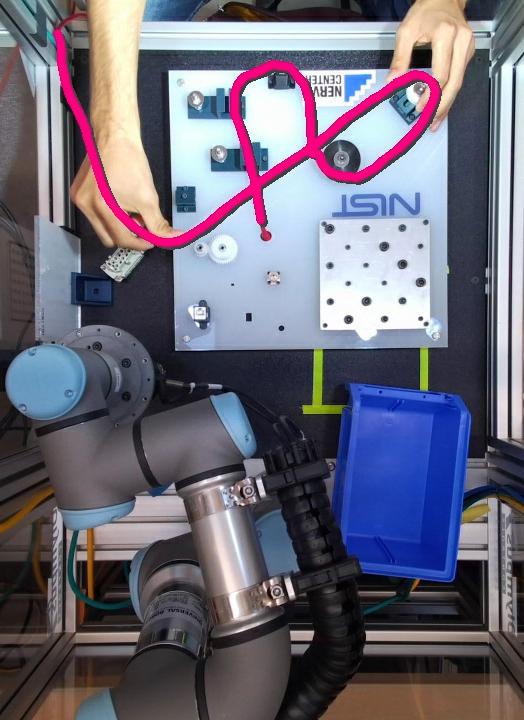}
    \end{subfigure}

    \medskip
    \begin{subfigure}[b]{0.48\textwidth}
        \includegraphics[width=0.32\textwidth]{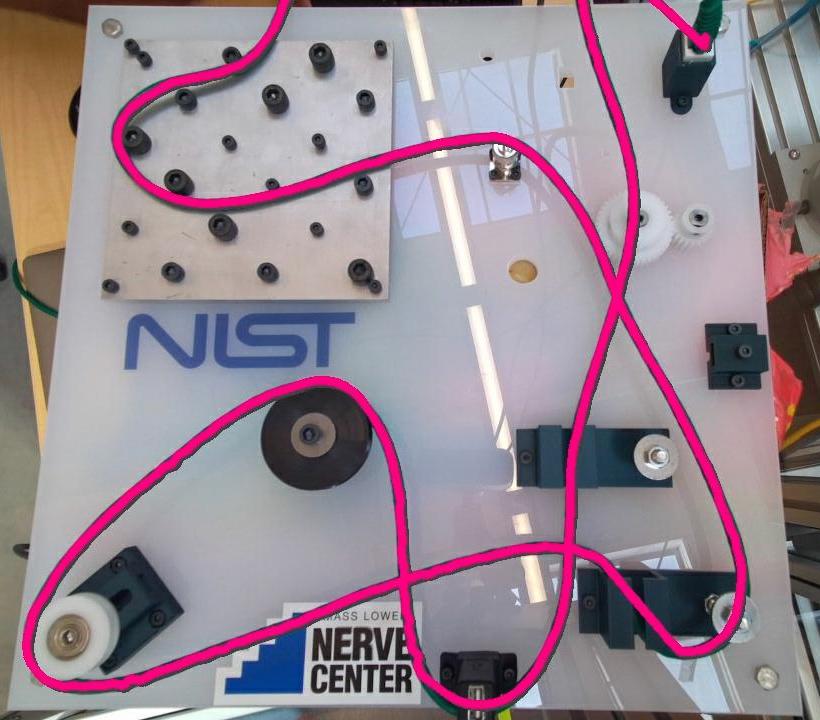}
        \hfill
        \includegraphics[width=0.32\textwidth]{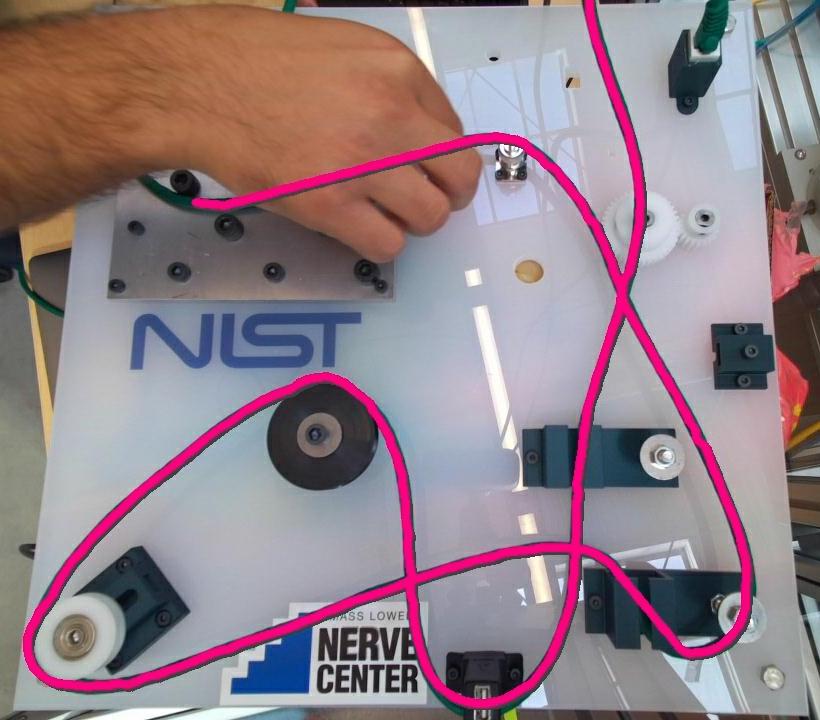}
        \hfill
        \includegraphics[width=0.32\textwidth]{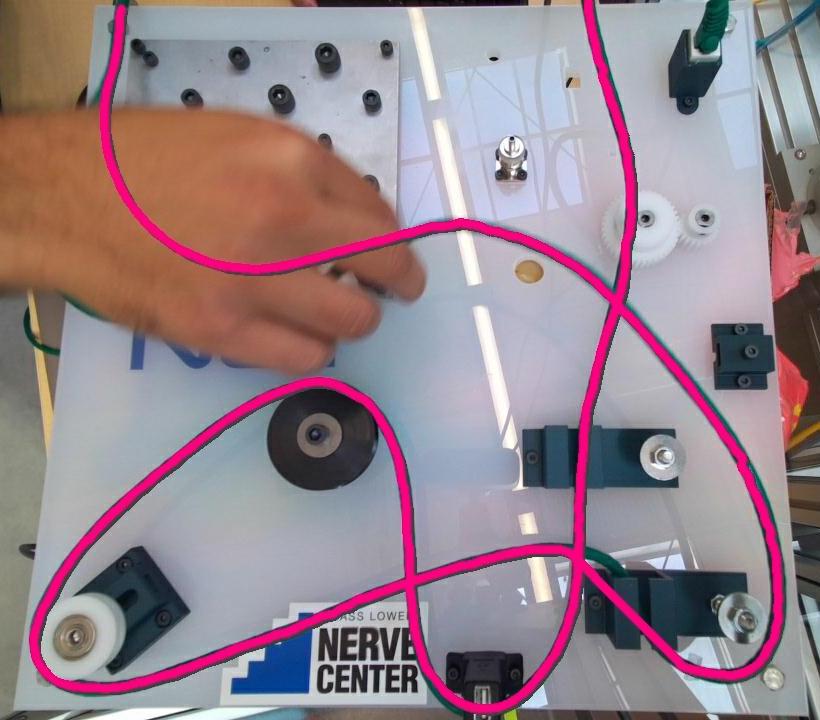}
    \end{subfigure}

    \medskip
    \begin{subfigure}[b]{0.48\textwidth}
        \includegraphics[width=0.23\textwidth, height=2cm]{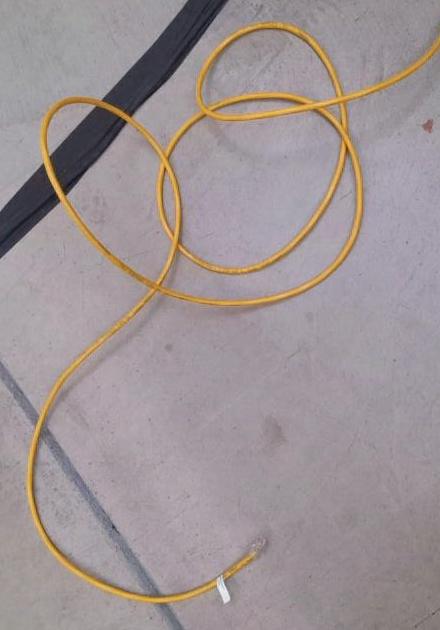}
        \hfill
        \includegraphics[width=0.23\textwidth, height=2cm]{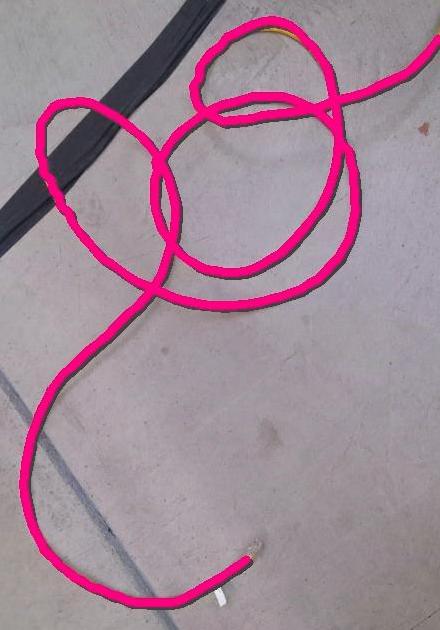}
        ~
        \includegraphics[width=0.23\textwidth, height=2cm]{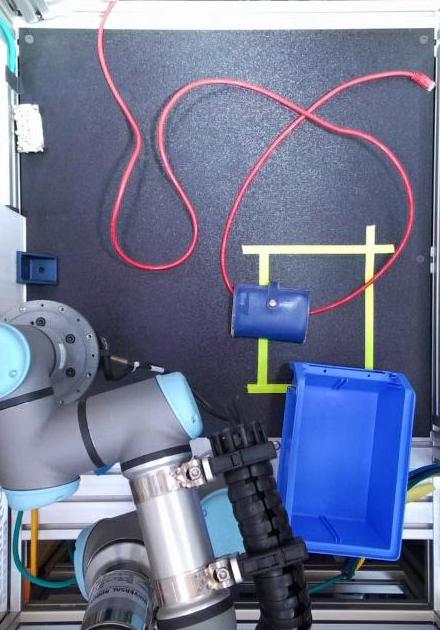}
        \hfill
        \includegraphics[width=0.23\textwidth, height=2cm]{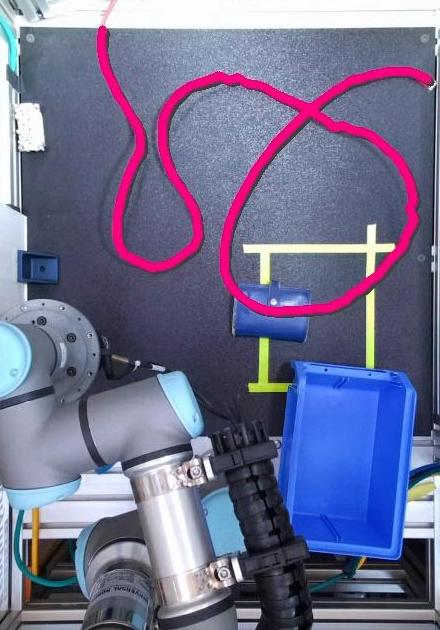}
    \end{subfigure}

\caption{Screenshots of example sequences with the overlaid detected cable (in magenta). The third row includes the original frame for comparison.}
\label{fig:results-3}
\end{figure}

We have used the method for cable routing and manipulation tasks~\cite{keipour2022icra}. The viability is tested on 7~video sequences with a total of 4,230~frames of size 1280$\times$720. Table~\ref{tbl:results} shows the quantitative results for the algorithm's accuracy on the whole cable in an image, for the occlusions filled, and for the merges performed. Mengyuan et al.~\cite{8972568} have used the root mean square of the Euclidean distance between their estimated and the ground-truth point positions on the DOO, which they reported as around 23~mm. Note that due to the lack of ground truth for the occluded areas and to focus on testing the key contributions of our proposed approach, we used a stricter measure that even when a single connection is incorrect, we counted the frame as incorrect detection. The occlusions are counted as incorrect when either a wrong connection is made or when the filled connection does not follow the actual cable's path. Finally, we noticed that the incorrect merges only rarely happen in places other than at occlusions, with only 148 cases, almost all of which happened at self-crossings. 

\begin{table}[!t]
\centering
\caption{Detection results on 7 video sequences.}
\label{tbl:results}
\begin{tabular}{|c|c|c|c|c|c|c|c|c|c|}
\hline
\rowcolor[HTML]{EFEFEF} 
~& Total & Correct & Incorrect & Accuracy \\ \hline
Frames         &  4,230 & 3,542 & 688 & 83.7\% \\ \hline
Occlusions & 26,456 & 23,991 & 2,465 & 90.7\% \\ \hline
Merges & 583,743 & 581,130 & 2,613 & 99.6\%\\ \hline
\end{tabular}
\end{table}

Our method's average detection time per frame across all the sequences is 0.537 seconds on a system with Intel® Core™ i9-10885H CPU and 64 GB DDR4 RAM. Figures~\ref{fig:results-1} and~\ref{fig:results-3} show snapshots of some video sequences and the detection results.

\section{Conclusions and Future Work} \label{sec:conclusion}

We presented a novel method for detecting deformable one-dimensional objects (e.g., ropes and cables) and showed the results. Our implementation is only 2-D, not tuned towards a specific condition, and is not optimally coded. Choices other than the weighted sum for the total cost function were not researched, and our selection of weights was not made optimally. Nevertheless, the results show promise with an almost 2~Hz detection rate on an HD image input. 

The method is flexible and can be tuned for specific 2-D and 3-D applications to provide near-perfect results. On the other hand, a more optimized implementation can take advantage of special data structures and parallelization to increase the method's speed by several orders of magnitude.

Note that it is not hard to find unstructured or adversarial situations with entanglements, occlusions, multiple close and parallel DOOs, and other complex scenarios that can easily confuse the proposed algorithm. This work aimed to provide a method that can assist in semi-structured situations rather than addressing those "crazy" scenarios.

The considerations for the 3-D case are provided for each step. However, we did not implement the 3-D case, and there may be unpredicted implementation challenges. In the future, the ideas of the method can be integrated with tracking methods to improve tracking accuracy. Its integration in a robotics pipeline can finally enable full autonomy in real-world robotics applications working with DOOs such as cables, surgical sutures, and ropes.

\addtolength{\textheight}{-0.0cm}   



\section*{Acknowledgment}

The authors would like to thank Olivier Pauly for great insights into approaching the problem from the medical imaging research point of view.


\bibliographystyle{IEEEtran}
\bibliography{paper-citations.bib}

\end{document}